\documentclass[preprint,5p,times,twocolumn,sort&compress]{elsarticle}

\usepackage{graphicx}
\usepackage{amssymb}

\usepackage{lineno}

\usepackage{booktabs}
\usepackage{amsmath}
\usepackage{bm}
\usepackage[normalem]{ulem}
\usepackage{textgreek}

\usepackage{xcolor}

\journal{Computers in Biology and Medicine}

\begin{document}

\begin{frontmatter}


\title{Rethinking multiscale cardiac electrophysiology with machine learning and predictive modelling}


\author[ecm,aero]{Chris~D.~Cantwell\corref{cor}}
\ead{c.cantwell@imperial.ac.uk}
\author[ecm,nhli]{Yumnah~Mohamied}
\author[ecm,nhli]{Konstantinos~N.~Tzortzis}
\author[ecm,bioeng]{Stef~Garasto}
\author[ecm,nhli]{Charles~Houston}
\author[ecm,nhli]{Rasheda~A.~Chowdhury}
\author[ecm,nhli]{Fu~Siong~Ng}
\author[ecm,bioeng]{Anil~A.~Bharath}
\author[ecm,nhli]{Nicholas~S.~Peters}


\address[ecm]{ElectroCardioMaths Group, Imperial College Centre for Cardiac Engineering, Imperial College London, London, UK}
\address[aero]{Department of Aeronautics, Imperial College London, South Kensington Campus, London, UK}
\address[nhli]{National Heart and Lung Institute, Imperial College London, South Kensington Campus, London, UK}
\address[bioeng]{Department of Bioengineering, Imperial College London, South Kensington Campus, London, UK}

\cortext[cor]{Corresponding author}

\begin{abstract}
We review some of the latest approaches to analysing cardiac 
electrophysiology data using machine learning and predictive modelling.
Cardiac arrhythmias, particularly atrial fibrillation, are a major global 
healthcare challenge. Treatment is often through catheter ablation, which 
involves the targeted localized destruction of regions of the myocardium 
responsible for initiating or perpetuating the arrhythmia. Ablation targets 
are either anatomically defined, or identified based on their functional 
properties as determined through the analysis of contact intracardiac 
electrograms acquired with increasing spatial density by modern electroanatomic 
mapping 
systems. While numerous quantitative approaches have been investigated over the 
past decades for identifying these critical curative sites, few have provided a 
reliable and reproducible advance in success rates.
Machine learning techniques, including recent deep-learning approaches, offer a 
potential route to gaining new insight from this wealth of highly complex 
spatio-temporal information that existing methods struggle to analyse. Coupled 
with predictive modelling, these techniques offer exciting opportunities to 
advance the field and produce more accurate diagnoses and robust personalised 
treatment.
We outline some of these methods and illustrate their use in making
predictions from the contact electrogram and augmenting predictive modelling 
tools, both by more rapidly predicting 
future states of the system and by inferring the parameters of these models 
from experimental observations.
\end{abstract}


\end{frontmatter}


\section{Introduction}
\label{s:intro}

Cardiac arrhythmias, particularly atrial fibrillation (AF), are a major global 
healthcare challenge in the developed world. Arrhythmias describe the abnormal 
and self-perpetuating propagation of action potentials (AP) within the 
myocardium. Their initiation and maintenance are incompletely understood and 
this has hindered the development of effective and reliable therapy. Treatment 
for AF is often through catheter ablation, where the regions of myocardium 
determined to be responsible for initiating or perpetuating the disturbance are 
targeted and made electrically inactive through the localised application of 
radio-frequency energy or freezing. For paroxysmal AF, catheter ablation 
delivers relatively good outcomes, with success rates in the region of 80-90 
percent \cite{phlips_2018}. However, outcomes of catheter ablation in patients 
with persistent AF remain disappointing, and is effective in only approximately 
50 percent of patients, despite all forms of adjunctive ablation strategies 
\cite{verma_2015}.

Identifying the critical sites responsible for abnormal AF maintenance has been 
a major focus of research, with a number of driving mechanisms, including 
rotors \cite{davidenko_1992}, multiple wavelets \cite{moe_1964} and epi-endo 
disassociation \cite{verheule_2014}, being proposed. Recent clinical studies, 
such as the CONFIRM study \cite{narayan_2012}, have tested the new approaches 
of catheter ablation by targeting the foci of rotational drivers, with 
initially 
promising results showing that 86\% of 101 cases achieved AF termination or 
slowing. However, subsequent studies suggest more moderate outcomes with 
Steinberg et al \cite{steinberg_2017} reporting only 4.7\% of 47 cases achieved 
AF termination, while 60\% documented recurrence within 12 months. The efficacy 
of this technique may be in part due to the poor spatial resolution of the 
global mapping catheter used \cite{roney_2017}.
Techniques involving the targeting of complex fractionated atrial electrograms 
(CFAE) \cite{nademanee_2010}, high dominant frequency (DF) 
\cite{latchamsetty_2009} and singularities identified during phase mapping 
\cite{nattel_2017} have each been used as strategies for terminating 
arrhythmias. However, none of these adjunctive ablation strategies have been 
shown to add any value to the conventional approach of electrically isolating 
the pulmonary veins \cite{verma_2015}. Part of the reason for this may be that 
they each discard a large proportion of the information content of the acquired
electrogram signals or their spatio-temporal association during analysis. 
Additionally, not all identified sites may be critical to the perpetuation of 
the arrhythmia, leading to excessive ablation. The complexity of the underlying 
electro-architecture of myocardium therefore requires a more sophisticated, 
personalised and multi-faceted approach to address the challenge of treating 
AF. 

The principle data modality used clinically for the treatment of AF is the 
contact electrogram, which arises from the superposition of electric fields 
induced by 
charged ions moving across cell membranes in the myocardium. It is the 
electrical signature of action potential propagation through tissue which 
implicitly encodes the functional and structural characteristics of the local 
substrate. The electrogram therefore provides a wealth of information which is 
rarely fully utilised in current clinical practice. Electrograms are normally 
only broadly categorised by binary descriptors -- such as \emph{simple} or 
\emph{complex}, \emph{early} or \emph{late} \cite{kanagaratnam_2008, 
    schilling_1998}, fractionated or non-fractionated -- with much of the 
signal 
content effectively discarded.
Despite a number of studies based on interpreting clinical electrogram data 
\cite{orozco_2016,duque_2017}, these do not investigate how electrogram 
morphology is influenced by individual electro-architectural factors. Our 
knowledge about the direct effects of electrical remodelling on electrogram 
morphology is consequently poor, considering the number of these abnormalities 
related to cardiac diseases \cite{kleber_2004}. Leveraging the electrogram to 
infer electroarchitectural properties of the myocardium may therefore provide 
new direct insight in locating critical sites for ablation.

Multiple concurrently recorded electrograms may be combined to evaluate the 
spatio-temporal propagation patterns occurring in the tissue. This activity can 
also be inferred from the surface of the body \cite{ghanem_2005}.
More recently, predictive modelling of action potential propagation is emerging 
as a potential tool for personalised testing and optimisation of interventions 
\cite{boyle_2018}, but this technology is heavily dependent on the accuracy of 
the underlying calibration of parameters. This can only be achieved by fully 
leveraging the huge wealth of information now available clinically. The data 
science revolution in the form of sophisticated machine learning algorithms and 
increasing availability of computing power, opens up possibilities to manage 
this data overload, both in terms of learning from the data, inferring model 
parameters and consequently making increasingly accurate predictions.

\subsection{Machine learning in cardiac electrophysiology}
Machine learning describes a class of algorithms which \emph{learn} model 
parameters from a set of training data (for which outcomes may, or may not, be 
known) with the purpose of accurately predicting outcomes for previously unseen 
data. Training data that includes associated outcome labels can be used for 
\emph{supervised} learning in which the algorithm uses this knowledge to 
directly improve its prediction. In contrast, \emph{unsupervised} learning 
seeks patterns in the data with more limited guidance, of which clustering is a 
common example. Although there is considerable overlap, machine learning 
methods are 
considered to differ from more conventional statistical modelling, such as 
regression, in that they are more concerned with the predictive accuracy of the 
resulting model rather than the ability to explain the reasoning behind its 
parameters and determining concrete relationships between the data. The high 
accuracy of some of the more recent machine learning methods -- which are 
virtually impossible to analyse analytically -- has justified this lack of 
transparency.

All machine learning algorithms seek some form of mapping that models the 
relationship 
between input data and outcome.
In an abstract context, we suppose that we have a model $f$, governed by one or 
more parameters $\bm{\theta}$, which maps an input $\mathbf{x}$ to some output 
$\mathbf{y}$, under the relation
\begin{align}
f(\mathbf{x},\bm{\theta}) = \mathbf{y}.
\label{e:ml-model}
\end{align}
The form and dimensions of $\mathbf{x}$ and $\mathbf{y}$ in 
Equation~\ref{e:ml-model} are a function of the particular problem under 
consideration. For example, $\mathbf{x}$ may be a large one-dimensional vector 
(time-series) in the case of a music-classification problem, or a 
two-dimensional image in the case of object recognition. For regressions, the 
output $\mathbf{y}$ may be a prediction of the dependent quantity, while for 
classification 
problems, $\mathbf{y}$ is usually a label which assigns the corresponding input 
to a single class. The size of $\bm{\theta}$ depends on both the problem and 
also on the chosen model. For example, for a 
linear regression between two variables $\bm{\theta}$ would consist of only two 
values (namely the \emph{slope} and \emph{intercept}), while for a many-layered 
deep neural network with high-dimensional input data, the size of $\bm{\theta}$ 
may be of $\mathcal{O}(10^6)$ or more.

Broadly speaking, the process of \emph{training} a supervised machine learning 
algorithm is 
the notion of seeking $\bm{\theta}$ such that, for some set of training input 
data $\{\mathbf{x}\}$ with corresponding outcomes $\{\mathbf{y}\}$, a given 
\emph{loss function} is minimised. While there are a number of loss functions, 
each with their own properties \cite{janocha_2017}, a simple loss function 
might be the $\mathcal{L}_1$ loss function which computes the differences 
between the predicted outputs and the actual outputs and is given by
\begin{align}
\epsilon = \sum_{i=1}^N ||f(\mathbf{x}_i,\bm{\theta}) - \mathbf{y}_i||_1.
\end{align}
If sufficient (and suitable) training data are used with an appropriate 
model, the expectation is that the model will then correctly predict the 
outcomes for other inputs which did not form part of these data.

Supervised machine learning is increasingly being used in medicine 
\cite{deo_2015}. One area of cardiac electrophysiology in which machine 
learning has become particularly 
prevalent to date is the analysis of the Electrocardiogram (ECG), in part due 
to its wide availability and its
potential to conveniently provide important information about cardiac function 
without intervention. There now exists a substantial body of literature on the 
application of machine learning tools to classify ECGs. A review of some of the 
earlier work is given by \cite{jambukia_2015}. The recent PhysioNet challenge 
to classify single-lead ECG segments into four categories (sinus rhythm, AF, 
other rhythm or too noisy) has catalysed developments in this area 
\cite{clifford_2017}. 
Most approaches require some form of preprocessing of the signal, including 
de-noising and correcting for baseline wander.
While convolutional neural networks \cite{hong_2017, zihlmann_2017, xiong_2017, 
rajpurkar_2017, rubin_2017, warrick_2017, kamaleswaran_2018, acharya_2017} 
and recurrent neural networks \cite{teijeiro_2017,hong_2017} are gaining 
popularity, many studies still achieved accurate classification results using 
other algorithms such as ensembles of decision trees (random forests) 
\cite{bin_2017, zabihi_2017}, multi-level binary classifiers \cite{datta_2017} 
and least-squares support vector machine classifiers \cite{billeci_2017}. The 
use of these approaches in combination also provides accurate classification 
\cite{zihlmann_2017, teijeiro_2017}. Recently, online real-time feature 
extraction and classification of ECGs using machine learning is being explored 
\cite{sutton_2018} and similar approaches are being used to diagnose more 
specific cardiac abnormalities \cite{sengupta_2018}.

In contrast, relatively little attention has been given to applying machine 
learning to make predictions from the contact intracardiac electrogram, or to 
predict the 
spatio-temporal patterns of activation in myocardium. Recently, there have been 
studies to characterise AF using \emph{in silico} or clinical contact
electrograms \cite{orozco_2016, duque_2017}, as well as for the automated 
location of \emph{in silico} re-entrant drivers using electrograms 
\cite{mcgillivray_2018}.

\subsection{Predictive numerical modelling}
Numerical modelling assumes the system under observation obeys particular 
physical laws, known \emph{a priori} and often represented in the form of 
partial differential equations, which are used to predict the future state of 
the system given an initial state. These equations often contain a number of 
parameters, which are estimated from experimental observations or experience.

While predictive modelling has advanced significantly in the field of cardiac 
electrophysiology for the past decade, only recently have the numerical methods 
and clinical imaging technologies improved sufficiently to allow 
viable predictions to be made on anatomically accurate geometries 
\cite{arevalo_2016,prakosa_2018}. However, challenges 
still remain in how to accurately personalise and validate these models, as 
well as how to safely incorporate them into clinical practice.

\subsection{Outline}
In this review, we describe some of the opportunities machine learning can 
provide in the field of cardiac electrophysiology. We illustrate these through 
examples as well as discuss their potential impact on arrhythmia management.
We begin with the contact electrogram – the data modality on which much of modern clinical electrophysiology is based. We introduce machine learning approaches to analysing and classifying these signals, contrasting both feature-based methods and deep neural networks, and show how they can be used to potentially elucidate a wealth of electro-architectural information about the myocardial substrate. We then discuss recent advances by our group in modelling action potential propagation and how machine learning might supplement and extend these methods to improve our ability to create personalised models which can be used on clinically relevant timescales.

\section{Feature-based classifiers}
\label{s:egm}
\emph{Features} describe characteristics of a process being observed. They are 
often represented in a numerical form and together form a feature vector. A 
feature-based machine learning algorithm then uses these feature vectors as 
input during both training and prediction. The selection of informative 
features is critical to the effectiveness of a machine learning algorithm to 
predict the correct output label. When 
a large number of features are available algorithms may struggle to generalise 
due to redundancy of information between features. Feature selection algorithms 
can alleviate this issue by selecting a subset of features which promote 
learning and improve the ability of the algorithm to make accurate predictions. 
The identification of which features are important in specific situations may 
also generate hypotheses to motivate further investigation of mechanistic links.

When the output label is one of a finite set of possible discrete values, the 
algorithm is termed a classifier. In the simplest case of binary 
classification, the accuracy of a learning algorithm may be characterised by a 
number of statistical measures. \emph{Sensitivity} describes the percentage of 
positive outcomes that are predicted as positive, 
while \emph{specificity} captures the proportion of negative outcomes predicted 
as negative. \emph{Positive (and negative) predictive value} instead captures 
the proportion of positive (and negative) predictions, which are truly 
positive (and negative).

A broad range of feature-based classifiers for supervised machine learning 
exist, and we refer the reader to previously published comprehensive reviews 
for specific details \cite{kotsiantis_2007,murthy_1998}. Both linear and 
non-linear classifiers map the input features to a set of classes $d$ using a 
weighted sum, 
\begin{align*}
d(\mathbf{x}) = f\left(\sum_i w_i (\phi(\mathbf{x}))_i\right)
\end{align*}
with the weights, $w_i$ learnt during training. In the linear case 
$\phi(\mathbf{x}) = \mathbf{x}$. The function $f$ then maps the result of the 
sum onto the different 
classes and may be a simple threshold function in the case of a binary 
classification, or probability densities more generally. While in general not 
as accurate as non-linear classifiers, linear classifiers are typically faster 
and so may be more effective in time-critical applications \cite{yuan_2012}.

Several approaches may be used to determine the weights of linear classifiers. 
Linear discriminant analysis \cite{riffenburgh_1957} seeks weights which best 
separate inputs $\mathbf{x}$ of different classes. Support vector machines 
(SVMs) \cite{burges_1998} instead seek to maximise the margin between a 
hyperplane and the two data classes it separates.
The $k$-nearest neighbour classifier is a non-linear classifier which uses the 
classes of the nearest training samples to predict the classification for 
unseen samples \cite{cover_1967}.
Decision trees \cite{murthy_1998} approach the classification problem 
feature-by-feature with branches in the tree representing different values a 
feature can assume. The leaves of the tree denote the final classification.

The performance of the above predictors can often be improved using the method 
of bootstrap aggregation, or \emph{bagging} \cite{breiman_1996}. Rather than 
training a single predictor on a training dataset, a number of training 
datasets are generated by drawing observations at random but with replacement 
and predictors are trained on each of these \emph{bootstrap} datasets. When 
making a prediction, the results of these predictors are \emph{aggregated} -- 
usually by voting when performing classification -- to form the final predictor.
Random forests \cite{breiman_2001} extend the bootstrap aggregation of decision 
trees by additionally 
selecting random subsets of features when deciding how to split at each node. 
These approaches help to overcome the problem of over-fitting often 
present with decision trees where they fail to generalise.

\subsection{Application to electrogram classification}
\label{s:egm-app}
We explore the use of supervised machine learning to classify individual 
electrograms based on the presence of cellular abnormalities.
For initial proof of concept we use signals acquired from cell monolayers in 
culture. While distinctive from clinical electrograms, they enable us to assess 
the capabilities of these algorithms in a controlled context and ensure signals 
can be labelled accurately.
We investigated the hypothesis that controlled functional modulations of the 
monolayers can be accurately and precisely predicted from the recorded unipolar 
electrogram morphology using supervised machine learning methods.
In particular, we sought the classification of electrical signals according to 
pharmacological gap junction uncoupling.

\subsubsection{Data acquisition and pre-processing}
\label{s:egm-data}
Electrogram recordings were acquired as previously described 
\cite{chowdhury_2018}. 
In brief, cell monolayers of neonatal rat ventricular myocytes (NRVMs) were 
seeded onto five microelectrode arrays (MEA), each consisting of 60 electrodes 
(MultiChannel Systems, Reutlingen, Germany). Ten-second recordings were made 
at a sampling frequency of 25kHz while pacing from one edge, before and 
after administration of 40\textmu M carbenoxolone (CBX) to increase gap 
junction 
uncoupling. No signal filtering was applied during data acquisition. All animal 
procedures were conducted according to the standards set by the EU Directive 
2010/63/EU.

\begin{figure}
    \centering
    \includegraphics[width=0.9\linewidth]{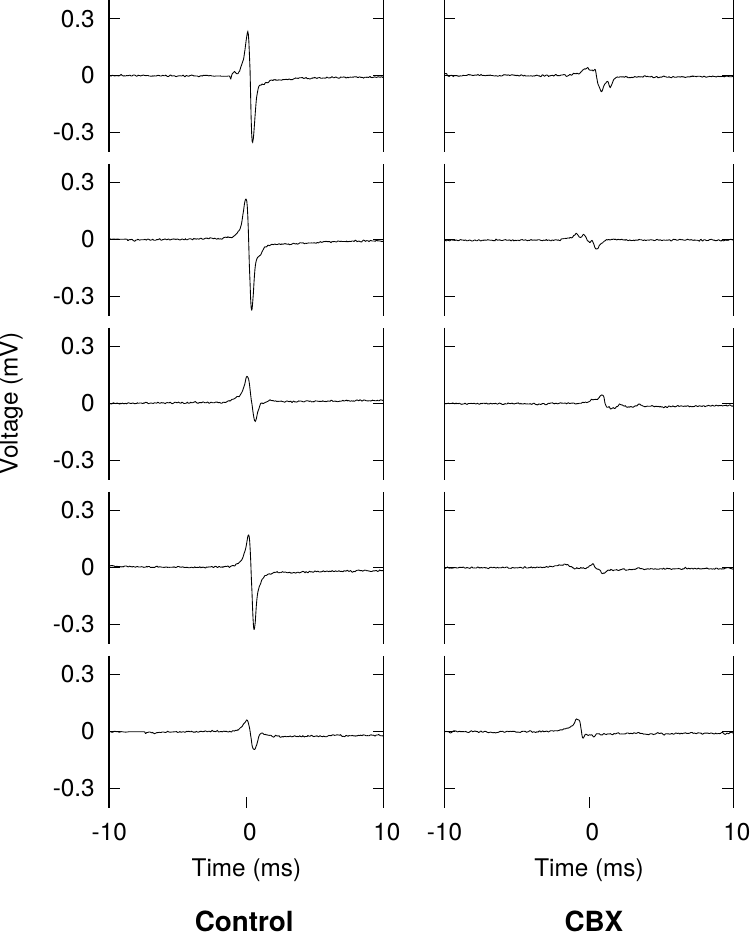}
    \caption{Examples of 20ms segments of electrogram recordings from the 
    Control and CBX groups, after removal of the stimulus artefact.}
    \label{f:egms}
\end{figure}

Pacing artefacts were removed by approximating the exponentially decaying 
stimulus deflection with a rational polynomial and subtracting it from the 
recorded signals. The dataset was further curated to remove recordings from 
electrodes where no further deflections were present. This resulted in 485 
control electrograms and 471 electrograms after treatment with CBX.
The dataset was partitioned into a training dataset and a testing dataset. The 
testing dataset consisted of all electrograms recorded from one plate (90 
control and 104 CBX electrograms). The training dataset consisted of the 
remaining four plates (395 control and 367 CBX electrograms). 
Examples of electrograms from the control and CBX classes are shown in 
Figure~\ref{f:egms}.

\subsubsection{Feature extraction}
\label{s:egm-feat}
The high sampling rate of the electrograms would result in very 
high-dimensional 
input data if used directly. We mapped each signal onto a 
pre-defined feature space of much lower dimension.
Each electrogram recording was represented by a fixed set of 27 time-, 
frequency- and morphological-based features, extracted from the signal using a 
custom-written algorithm (Matlab R2017b). Details of these features are 
provided in the Supplementary Material. Sequential Forward Selection 
(SFS) \cite{whitney_1971} was used to select a subset of these features which 
were sufficient to differentiate the control and CBX classes. In brief, SFS is 
a \emph{bottom-up} approach to choosing discriminatory features. Starting with 
an empty \emph{feature set}, the algorithm sequentially adds features 
from the candidate set of features which maximises a given objective function, 
until the addition of further features provides no improvement. Classification 
accuracy is used as the objective function. Feature 
selection indicated that only three of the 27 features considered were 
sufficient to distinguish the control and CBX classes: \emph{electrogram 
amplitude}, \emph{standard deviation of the autocorrelation function} and the 
\emph{scale with minimum energy} in the continuous wavelet transform of the 
signal.
The set of values from the selected electrogram characteristics form the 
feature vector for that electrogram. These feature vectors were 
subsequently used to train the classifier.

The bootstrap aggregating (or \emph{bagging}) ensemble tree method 
\cite{breiman_1996} was applied during both feature selection and 
classification training.

\subsubsection{Results}

\begin{figure}[tb]
    \centering
    \includegraphics[width=\linewidth]{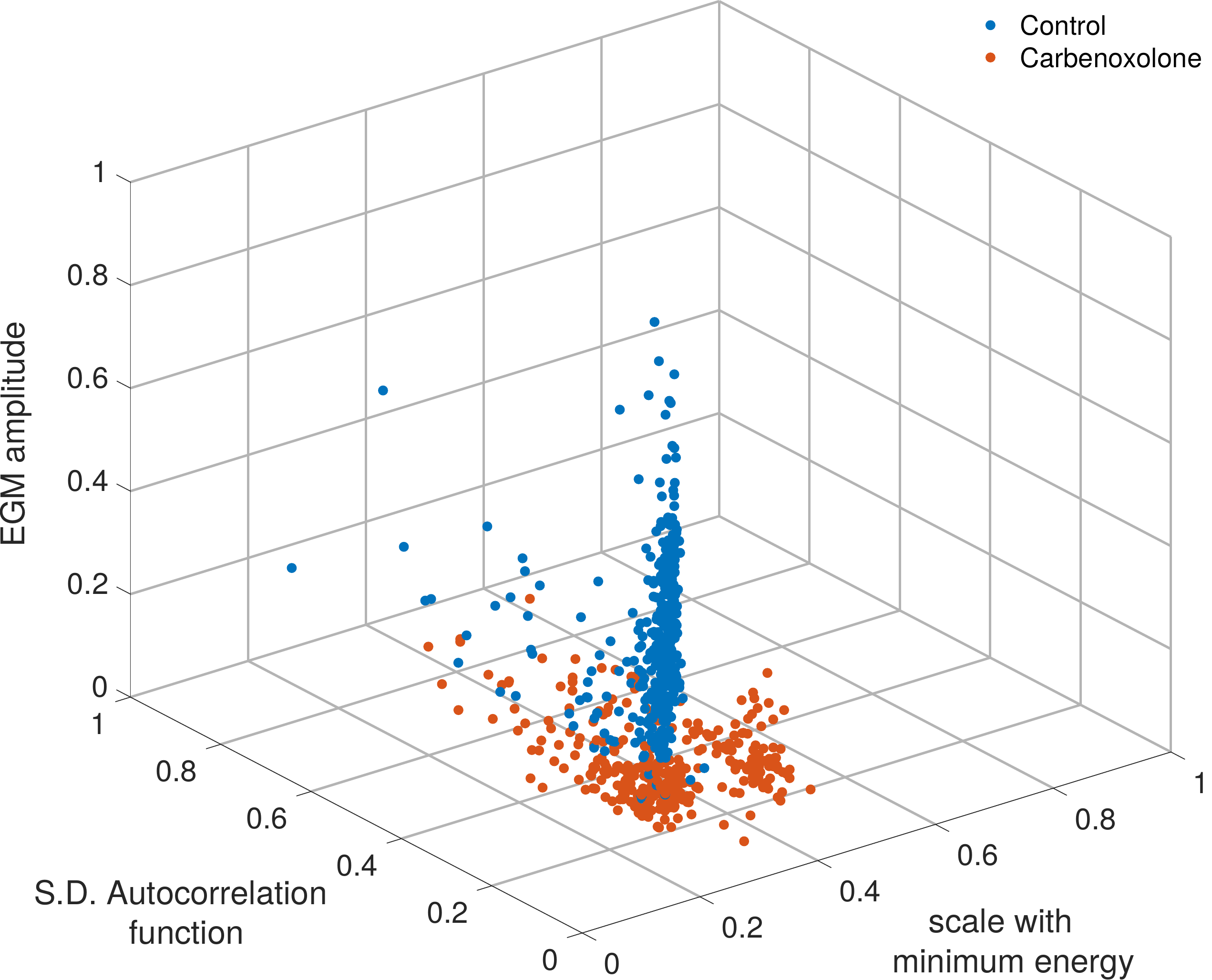}\\
    \caption{3D scatter plot of the most relevant features, normalised in the 
    interval [0,1], as determined by SFS. No single feature clearly 
    discriminates between the control and carbenoxolone classes.}
    \label{f:egm-feat-results}
\end{figure}

Figure~\ref{f:egm-feat-results} shows the distribution of all electrogram 
feature vectors from 
the training dataset in the corresponding three-dimensional feature space. It 
is evident that no single feature alone effectively discriminates between the 
two classes.

\begin{figure}[tb]
    \centering
    \includegraphics[width=0.6\linewidth]{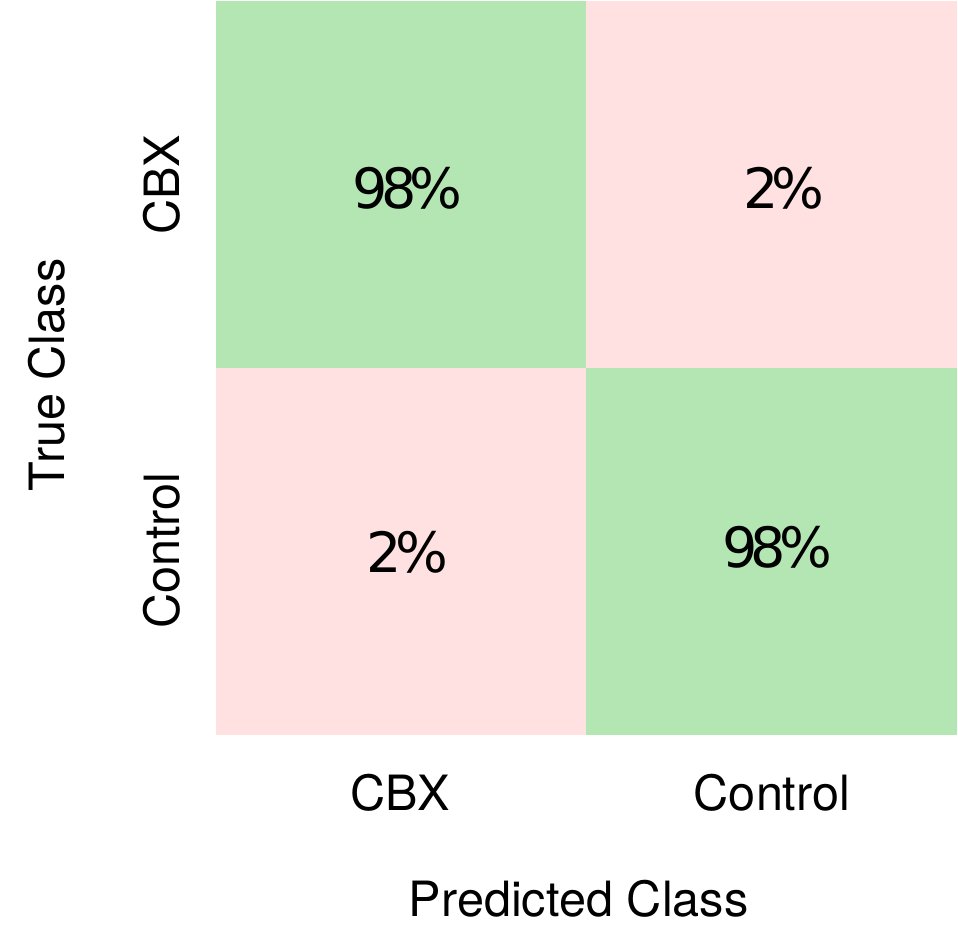}
    \caption{Confusion matrix comparing the predictability of classes using the 
    training dataset. Diagonal cells (green) show the percentage of 
    electrograms that were correctly classified.}
    \label{f:egm-feat-confusion}
\end{figure}

Validation was performed on the training data using ten-fold cross-validation.
A total of 30 decision trees, with a leaf size of one, were used for the 
bagging ensemble method.
The performance characteristics of the trained model are given in 
Table~\ref{t:egm-feat-results}. A specificity of 98.1\% was achieved on the 
training 
data when using only the three features chosen by SFS. This is illustrated by 
the \emph{confusion matrix} shown in Figure~\ref{f:egm-feat-confusion}, which 
compares predicted class against true class.
This indicates the model 
was capable of accurately distinguishing the classes. The model performance was 
then measured using the unseen test dataset of 194 electrograms. It achieved a 
96.7\% sensitivity, 84.4\% specificity, 83.8\% positive predictive value and 
96.8\% negative predictive value, indicating the model has generalised 
successfully.

\begin{table*}[tb]
    \centering
    \begin{tabular}{p{5cm}p{5cm}p{5cm}}
        \toprule
        & Classification training \newline (734 EGMs) 
        & Prediction model testing \newline (194 EGMs) \\
        \midrule
        Sensitivity & 98.1\% & 96.7\% \\
        Specificity & 98.3\% & 84.4\% \\
        Positive predictive value & 98.4\% & 83.8\% \\
        Negative predictive value & 97.7\% & 96.8\% \\
        Error rate & 1.9\% & 10\% \\
        \bottomrule
    \end{tabular}
    \caption{Performance of classification training using the Bagging Ensemble 
    method and evaluation of the subsequent prediction model on the test 
    dataset.}
    \label{t:egm-feat-results}
\end{table*}

The computational time of the training process was also measured. Calculation 
of the 
electrogram features was the dominant cost, at 18.84 $\pm$ 5.2 seconds per 
electrogram. The time taken for the bagging ensemble method to train a model 
using the feature vectors was 3.25 $\pm$ 0.13 seconds.

\section{Convolutional Neural Networks}
\label{s:cnn}

One of the limitations of conventional machine-learning techniques is their 
inability to be applied directly to high-dimensional data, necessitating a 
transformation into a suitable feature-space representation that captures 
characteristics of 
the data relevant for discrimination, while remaining invariant to irrelevant 
aspects.
As illustrated in Section~\ref{s:egm-feat}, extracting these features often 
requires significant domain-specific knowledge and expertise 
in order to hand-engineer suitable algorithms, and thus produce an informative 
representation that supports discrimination. This process can, however, be 
circumvented if the feature-extraction step is automated.
 
Representation learning \cite{bengio_2013} encompasses a set of techniques 
within the field of 
machine learning which enable a model to automatically learn and discover for 
itself discriminating features directly from the raw observational data. Among 
the most popular of current techniques are layered artificial neural networks, 
which take inspiration from neuroscience.
They are composed of artificial neurons (simplified versions of biological 
neurons) arranged in layers, where the neurons in one layer are connected 
to many, if not all, of the neurons in the subsequent layer. The artificial 
neuron is a non-linear mapping from an input value to an output value. The 
output values from all the neurons in one layer are each multiplied by 
adjustable parameters, called weights, to form a weighted sum as input to one 
individual neuron in the subsequent layer. A neural network is, therefore, a 
complex system of weighted non-linear functions nested within each other and it 
is these weights that must be learnt in order for the network to accurately map 
raw input data to a desired output.

Neural networks learn their own weights during training: initially the weight 
values are randomly selected and thus when a model is given input it is 
initially highly unlikely to predict the correct output label. During training, 
the model is shown raw input data and the associated label or output value. For 
each input example the model makes a prediction based on the current weight 
values, and the error between the prediction and the true desired output is 
measured. The weights are then modified in order to minimise this error via a 
process called back-propagation, the details of which are provided in 
\cite{lecun_2015}. An \emph{epoch} is defined as a complete pass over the 
training data. Unlike the discriminant classifier of Section~\ref{s:egm-app} 
which only needs one pass, neural networks benefit from multiple passes over 
the training data. With sufficient training data and sufficient iterations 
(epochs) over all the data the weights converge onto values that enable 
the model to make accurate predictions for the training dataset. The trained 
model is subsequently tested on a validation dataset it has never before seen 
in order to measure its predictive power.

The weights of a neural network can be seen as the features of the learnt 
representation, automatically discovered without the need for manually-designed 
feature detectors. When the network contains several layers in between the 
input layer and the final output layer -- known as \emph{hidden} layers -- it 
is referred to as a deep network, from which the term deep-learning arises. 
These models are therefore representation-learning methods with multiple 
non-linear layers, each transforming the representation, beginning with the raw 
input, into increasingly more discriminative representations.

Today, deep-learning techniques provide state-of-the-art solutions in the 
fields of object recognition \cite{krizhevsky_2012, szegedy_2015} and detection 
\cite{ren_2017}, speech recognition \cite{hinton_2012} and natural language 
processing \cite{sutskever_2014, collobert_2011}, and are increasingly being 
used in other domains such as genomics \cite{leung_2014} and challenging 
segmentation problems required for geometric reconstruction in biomedical 
imaging \cite{helmstaedter_2013}. Recently, studies 
have applied deep neural networks to the ECG signal~\cite{hong_2017, 
zihlmann_2017, xiong_2017, rajpurkar_2017, rubin_2017, warrick_2017, 
kamaleswaran_2018, acharya_2017}. These studies 
have all made use of convolutional neural networks (or 
\emph{convnets}).

Fully-connected neural networks treat 
neighbouring data points identically to those spaced far apart. In the case of 
time-series data, they 
do not account for the temporal structure and autocorrelation that may be 
present in the raw input data. As such, they may fail to recognise, for 
example, a QRS complex in an ECG if it had been shifted in time by half a beat 
compared to the training data examples. Convnets are deep-learning 
architectures that cater to this need for translation-invariance. 
They exploit compositional hierarchies -- whereby higher-level features are 
generated by accumulating a set of lower-level features -- often exhibited in 
datasets derived from the natural world.

\subsection{Application to classifying electrograms}

\begin{figure}[tb]
    \centering
    \includegraphics[width=0.7\linewidth]{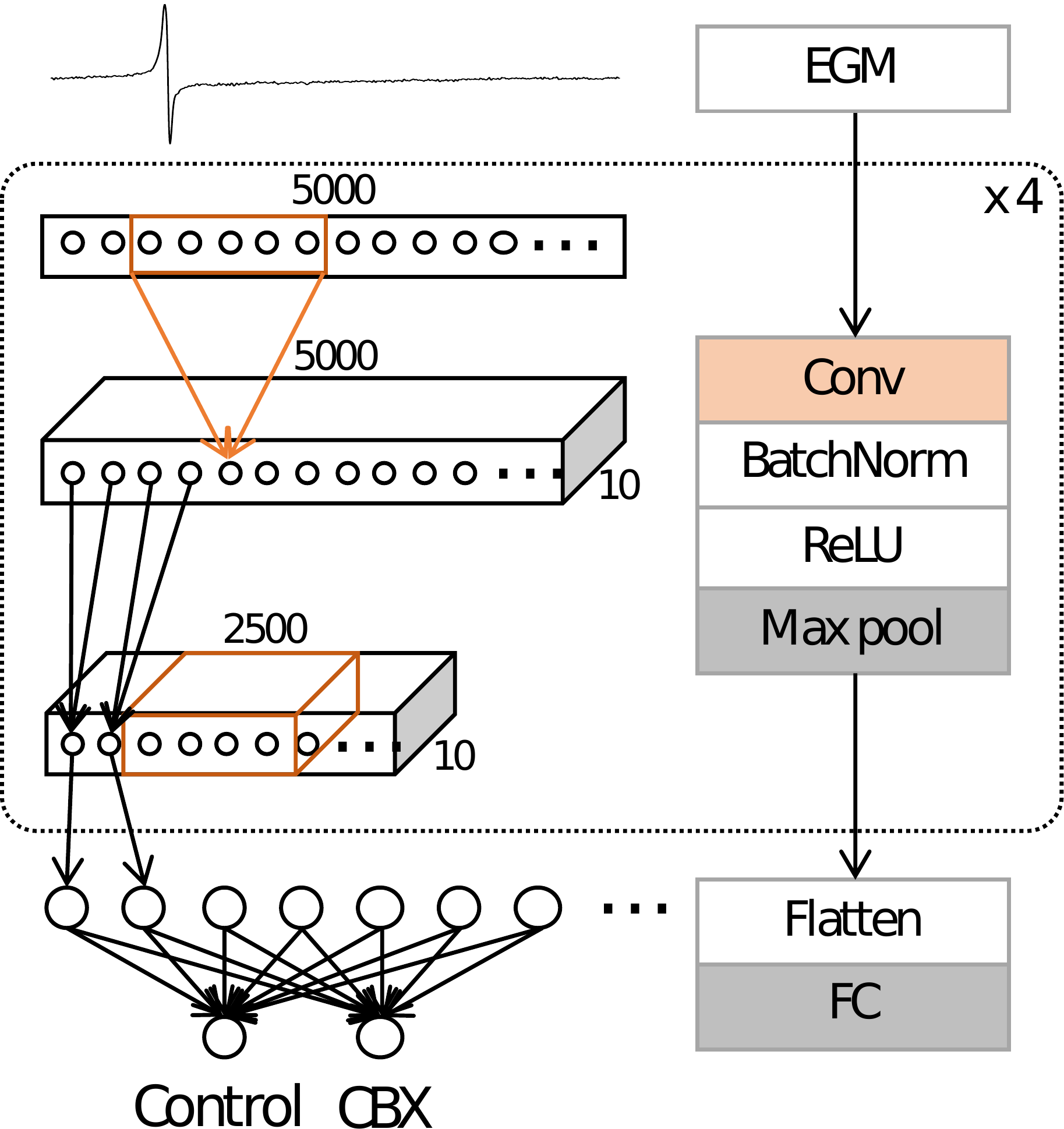}
    \caption{Schematic of the convolutional neural network.}
    \label{f:egm-dnn-arch}
\end{figure}

We demonstrate the application of convolutional 
neural networks, using the same data as described in Section~\ref{s:egm-data}, 
to perform the classification directly from the labelled time-series data.
The network is composed of four repeated blocks, each 
itself consisting of a convolutional layer, a batch normalisation layer 
\cite{ioffe_2015}, a non-linear ReLU layer \cite{nair_2010} and a max pooling 
layer \cite{goodfellow_2016}. Batch normalisation adjusts the inputs to a layer 
to have mean zero and standard deviation of one, and is a technique for 
improving the stability of the network. The ReLU layer is a form of activation 
function, while max pooling layers are used to down-sample the input as it 
progresses through the network. The last block is followed by one 
fully-connected layer to the two output classes. A schematic of the 
architecture is shown in Figure~\ref{f:egm-dnn-arch}.
 
\subsubsection{Training}
During training, a randomly selected one-second segment (thereby guaranteed to 
contain one deflection) was taken from the full ten-second recording and 
down-sampled to 5kHz. Training was carried out for 250 epochs using the Adam 
optimiser \cite{kingma_2014} with a variable learning rate starting at 0.0001 
and a weight-decay of zero. The error was evaluated using the cross-entropy 
loss function \cite{janocha_2017}, and training was repeated five times each 
with different random 
initial weights to evaluate the performance (averages and standard deviations), 
giving a measure 
of the robustness of the architecture and optimisation process. This was 
carried out using the PyTorch framework~\cite{paszke_2017} on an GTX 1080ti GPU 
(NVIDIA Corporation, USA).

Cross-validation was carried out to measure the robustness of the prediction 
model as well as to aid tuning of the parameters of the learning algorithms and 
design choices of the architecture (e.g. number of layers). Once these
hyper-parameters were sufficiently tuned, the validation and training data were 
combined (four plates) and the model retrained. It was then evaluated using the 
test data (one plate).
Trained models were evaluated by splitting the ten-second recordings from the 
test or validation datasets into ten segments corresponding to the ten 
deflections and classifying each segment. If six or more deflections were 
classified correctly, the recording was considered successfully classified and 
the uncertainty of this positive classification was computed using the binary 
entropy function. In practice, there was little 
variation between the ten deflections within a signal, resulting in consistent 
classification 
of each recording.

\subsubsection{Results}
\begin{table}[tb]
    \centering
    \begin{tabular}{p{2.5cm}p{1cm}p{4cm}}
    \toprule
 & Fold & Classification accuracy (\%) \\
    \midrule
Cross-validation & 1 & 96.7 $\pm$ 0.8 \\
                 & 2 & 97.2 $\pm$ 0.6 \\
                 & 3 & 96.7 $\pm$ 1.0 \\
                 & 4 & 94.0 $\pm$ 0.7 \\
\textbf{Testing}  && \textbf{96.3 $\pm$ 0.7} \\
    \bottomrule
    \end{tabular}
    \caption{Total classification accuracy results from the 4-fold 
    cross-validation and final testing. For each model, the convolutional 
    neural network was trained a total of five times to ensure the model was 
    robust to differences in random initialisations. Averages and standard 
    deviations of the classification accuracy are presented.}
    \label{t:egm-dnn-results}
\end{table}

\begin{figure}
\begin{center}
\includegraphics[width=0.6\linewidth]{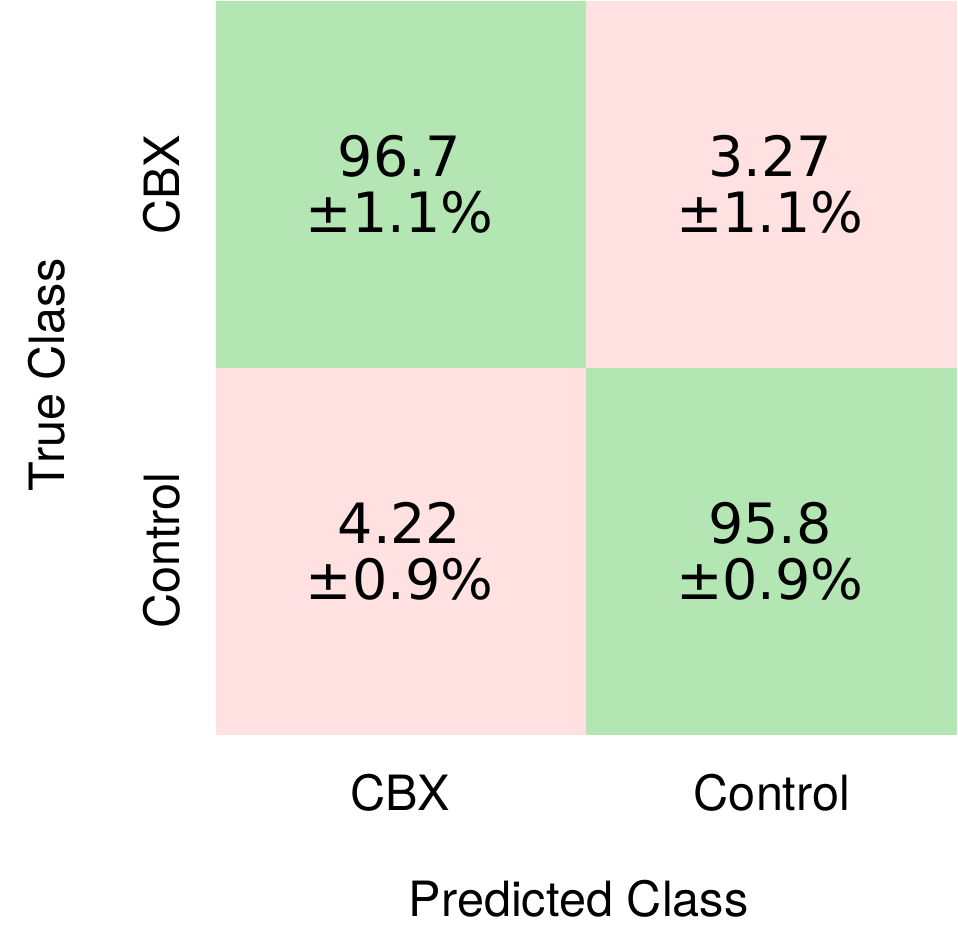}
\end{center}
\caption{Confusion matrix from the binary classification of electrograms before 
and after administration of carbenoxolone to monolayers of cultured myocytes.}
\label{f:egm-dnn-confusion}
\end{figure}

Table~\ref{t:egm-dnn-results} shows the classification accuracy 
using four-fold cross-validation after all design choices were made. The 
average accuracy across the folds was 96.2\% with a standard deviation of 
1.5\%, indicating a reliably robust model, and in general the entropy values 
remained low at approximately 0.05, indicating that each ten-second segment was 
consistently classified correctly or incorrectly. The standard deviation of 
the five repeats for each fold showed the models were converging to a 
similar optimum state regardless of the initial weight values. The final 
evaluation of the model on the test data resulted in an overall accuracy of 
96.3 $\pm$ 0.7\%, with results of 96.7 $\pm$ 1.1\% sensitivity, 95.8 $\pm$ 
0.9\% specificity, 96.4 $\pm$ 0.7\% positive prediction value and 96.2 $\pm$ 
1.2\% negative predictive value. The confusion matrix for this binary 
classification problem is shown in Figure~\ref{f:egm-dnn-confusion}. 

\section{Numerical Modelling}
Numerical modelling predicts the future behaviour of a dynamical system from a 
known state under an \emph{a priori} belief in the physical laws governing the 
system. In particular, it can allow observations to be extrapolated forward in 
time to help better understand the likely behaviour of a physical system and, 
by modulating the parameters and initial condition of the system appropriately, 
allow hypothetical scenarios to be explored \emph{in silico}. This makes it a 
potentially invaluable tool for both improving our understanding of the 
mechanisms driving arrhythmogenesis and as a direct clinical tool for aiding 
diagnosis and planning intervention.

\subsection{Tissue-scale continuum modelling}
The physical processes responsible for the cardiac action potential in a 
cardiac myocyte are a complex choreography of ion movements across the cell 
membrane. Mathematically, these are often described by systems of ordinary 
differential equations (ODEs) which range in complexity from two equations to 
more than twenty \cite{keener_2009}. The parameters of these equations have 
been chosen based on fitting the individual equations to experimental 
measurements. Cells are electrically coupled through gap junctions, which can 
be mathematically modelled as resistors. However, modelling the whole heart at 
a cellular scale is computationally intractable. Consequently, homogenisation 
of the discrete cell model leads to a bi-domain continuum model in the form of 
two partial differential equations (PDEs),
\begin{align}
    \nabla \cdot (\bm{\sigma}_i \nabla v) + \nabla \cdot (\bm{\sigma}_i 
    \nabla 
    v_e)
     &= \chi \left(C_m \frac{\partial v}{\partial t} + I_{ion}\right),\\
    \nabla \cdot (\bm{\sigma}_i \nabla v) + \nabla \cdot ((\bm{\sigma}_i + 
    \bm{\sigma}_e) \nabla v_e)
     &= 0,
\end{align}
supplemented with appropriate boundary conditions \cite{clayton_2011}. Here $v$ 
is the transmembrane potential, $v_e$ is the extracellular potential, $C_m$ is 
the membrane capacitance per unit area, $\chi$ is the cellular 
surface-to-volume ratio and 
$I_{ion}$ is the transmembrane current density from the coupled action 
potential ODE model. Anisotropy and 
heterogeneity of the myocardium is captured in the intracellular conductivity 
tensor $\bm{\sigma}_i$ and extracellular conductivity tensor $\bm{\sigma}_e$. 
With 
the further assumption of equal anisotropy ratios in these spaces, such that 
$\bm{\sigma}_i = \lambda \bm{\sigma}_e$, this system of equations can be 
further reduced to a single PDE, again with appropriate boundary conditions, 
known as the monodomain model,
\begin{align}
    \frac{\lambda}{1+\lambda}\nabla\cdot(\bm{\sigma}_i\nabla v)
    &= \chi \left(C_m \frac{\partial v}{\partial t} + I_{ion}\right)
    \label{e:monodomain}
\end{align}
A comprehensive review of the mathematical models used in the cardiac 
electrophysiology domain is given by Clayton \emph{et al} \cite{clayton_2011}. 
The monodomain model belongs to the class of reaction-diffusion PDEs. The 
diffusion component can be considered to relate to the biophysical process of 
ion propagation through gap junctions between cells, while the reaction 
component is the cumulative result of the action potential model describing the 
opening and closing of ion channels in the membrane (gating variables) and 
related ion movements into, or out of, the cell.

\subsection{Numerical methods for action potential propagation}
To solve equation~(\ref{e:monodomain}) in all but the most trivial of scenarios 
requires the use of numerical approximations. The continuous PDE is transformed 
into a system of algebraic equations which are more amenable to solution on a 
computer. This transformation may use one of several discretisation techniques, 
such as finite difference, finite element or spectral approximations and with 
sufficient spatial and temporal resolution can provide very accurate 
approximations to the true solution of the PDE. However, the wide range of 
time-scales on which the different physical processes in the model occur makes 
the numerical solution of this system challenging and may lead to long 
simulation times even with considerable computational resources.

One technique being explored within our group for modelling 
electrophysiology is the spectral/$hp$ element method \cite{cantwell_2015}. 
This approach combines the flexibility of the finite element method to model, 
for example, the complex geometry of the heart chambers, with the numerical 
benefit of spectral methods, by enriching the polynomial space of each element 
with higher-order basis functions. In particular, this allows an approximation 
of comparable accuracy to a conventional finite element discretisation to be 
achieved with a smaller algebraic system of equations, resulting in faster 
simulations and ultimately a shorter time to solution. We have also explored 
the simulation of action potential propagation in the left atrium using a 
surface representation of the chamber wall \cite{cantwell_2014}, further 
reducing the size of the numerical problem to solve.

Even with these advancements, the computational cost of using numerical methods 
to accurately perform 
predictive modelling is still substantial, with the time-to-solution being 
orders of magnitude higher than what might be required for an interactive 
clinical tool. Furthermore, the inference of model parameters is highly 
challenging and even more computationally costly. In Section~\ref{s:rnn} and 
Section~\ref{s:inference} we consider opportunities for machine learning 
techniques to complement modelling and help address these difficulties.

\section{Recurrent Neural Networks}
\label{s:rnn}
Deep neural networks can be trained to predict the future behaviour of 
a dynamical system \cite{tompson_2016, de_2017, ehrhardt_2017}, and 
subsequently their internal representation can be used to estimate the latent 
parameters of the system. Neural networks have been shown 
to be faster (at inference time) than commonly used numerical simulation 
approaches \cite{tompson_2016, de_2017}. Indeed, while they require large 
amounts of data to train, once the optimal network weights have been found, 
obtaining predictions from unseen inputs requires only a fraction of the time 
and computational resources in comparison to conventional numerical methods. 
Despite the drawback of generating only approximate solutions, leveraging the 
fast prediction performance of neural networks may enable large numbers of 
\emph{what-if} scenarios to be rapidly explored in a fraction of the time of 
conventional numerical modelling. Clinically, this might allow the viability of 
potential therapeutic strategies to be quickly tested and accelerate the 
calibration of more precise numerical modelling which can be used to further 
optimise the treatment. 

Recurrent neural networks differ from purely feed-forward neural networks, such 
as the convolution neural networks considered in Section~\ref{s:cnn}, in that 
they are designed for processing sequences of inputs. They incorporate a 
feedback loop where the output of each step in the sequence is added to the 
input of the next 
step \cite{goodfellow_2016}. These types of networks are therefore
well suited to dealing with data sequences. This allows 
information to be propagated along the sequence: each output will then be 
conditioned not only by the current input in the sequence, but also by all 
previous inputs. However, in practice, vanilla recurrent neural networks can 
only store information for a short number of steps.
Long short-term memory (LSTMs) networks are a 
particular variation of recurrent neural networks developed to alleviate this 
problem and allow the learning of longer term dependencies 
\cite{goodfellow_2016}. 

\subsection{Application to predicting two-dimensional diffusion}
\label{s:pred-dnn}
Here, we present a proof of principle study where we apply this approach to a 
two-dimensional diffusion problem with a spatially heterogeneous and 
anisotropic diffusion tensor. Diffusion is a key component of models of 
excitable media, such as cardiac tissues \cite{clayton_2011}. Our system is 
governed by the following equations:
\begin{align}
\frac{\partial v(x,y,t)}{\partial t} &=  \nabla(\mathbf{D}(x,y)\nabla 
v(x,y,t)), 
                && (x,y,t)\in\Omega\times [0,T], \\
v(x,y,0)   &= v_0(x,y) && (x,y) \in\Omega, \\
v(x,y,t)   &= 0 && (x,y,t)\in\partial\Omega\times [0,T].
\end{align}
Here, $\mathbf{D}$ is a diagonal 2x2 matrix with non-zero diagonal elements 
$d_0$ and $d_1$, governing diffusion on the horizontal and vertical axis, 
respectively. The computational domain $\Omega=[-2,2]^2$.

\subsubsection{Training data generation}
Numerical simulations were performed using a regular mesh of $80\times 80$ 
square elements, using a modified Legendre polynomial basis with polynomials up 
to order $P=5$. The solution over time was subsequently sampled on a regular 
$64\times 64$ grid for input into the neural network.
A total of 1600 numerical simulations were performed using Nektar++ 
\cite{cantwell_2015} 
with initial condition and diffusion fields drawn at random from a predefined 
distribution. The initial condition consisted of spatially smoothed noise. This 
was created by first generating a random spatial frequency spectrum in the 
Fourier domain. This had a two-dimensional frequency profile drop-off as 
$(f_x^2+f_y^2+f_c)^{\alpha/2}$, with alpha randomly chosen to be either $-1$ or 
$-2$ and $f_c=3$. The quantities $f_x$ and $f_y$ are measured in terms of 
cycles per domain length. Furthermore, there is a sharp cut-off at 
$f_x^2+f_y^2<f_o^2$, with $f_0=8,12$ or $16$ which is randomly drawn 
independently for each simulation. The spatial frequency profile was multiplied 
by standard Gaussian noise and phases were drawn from a uniform distribution 
between 0 and 2$\pi$. Finally, a symmetric version of the spectrum is inverted 
to obtain the initial condition in the spatial domain and then normalised to 
achieve a certain level of contrast. All initial conditions are gradually 
smoothed to zero when approaching the boundary of the domain. Moreover, they 
were first generated at a resolution of 128x128 pixels and then interpolated 
onto the mesh used for the simulation.

The diffusion field was characterised by six parameters. A line with random 
orientation ($\theta$) and location ($\beta$) is chosen to partition the 
domain, with one part denoted as healthy and the other as scarred tissue. The 
former is given a higher diffusion coefficient with respect to the latter. 
Diffusivity in the domain is therefore characterised by four different 
diffusion parameters: $d_0$, $d_1$, $d_{0,scar}$, $d_{1,scar}$, which are 
chosen to satisfy the conditions,
\begin{align}
\frac{\max(d_0,d_1)}{\min(d_0,d_1)}
&= \frac{\max(d_{0,scar},d_{1,scar})}{\min(d_{0,scar}, d_{1,scar})}
\equiv \gamma, \\
\frac{d_0}{d_{0,scar}} &= \frac{d_1}{d_{1,scar}} \equiv \lambda.
\end{align}
Here, the anisotropy ratio $\gamma$ and heterogeneity ratio $\lambda$ are 
randomly drawn from a uniform distribution on the intervals [1,3] and [2,7], 
respectively. Finally, for each simulation we randomly selected which of $d_0$ 
and $d_1$ was assigned to have the highest magnitude. This consequently 
determines the direction along which diffusion is fastest. The direction of 
fastest diffusion was assigned a value drawn from the interval [3.2, 3.8].

\subsubsection{Network architecture}
\begin{figure*}[tb]
    \centering
    \includegraphics[width=\linewidth]{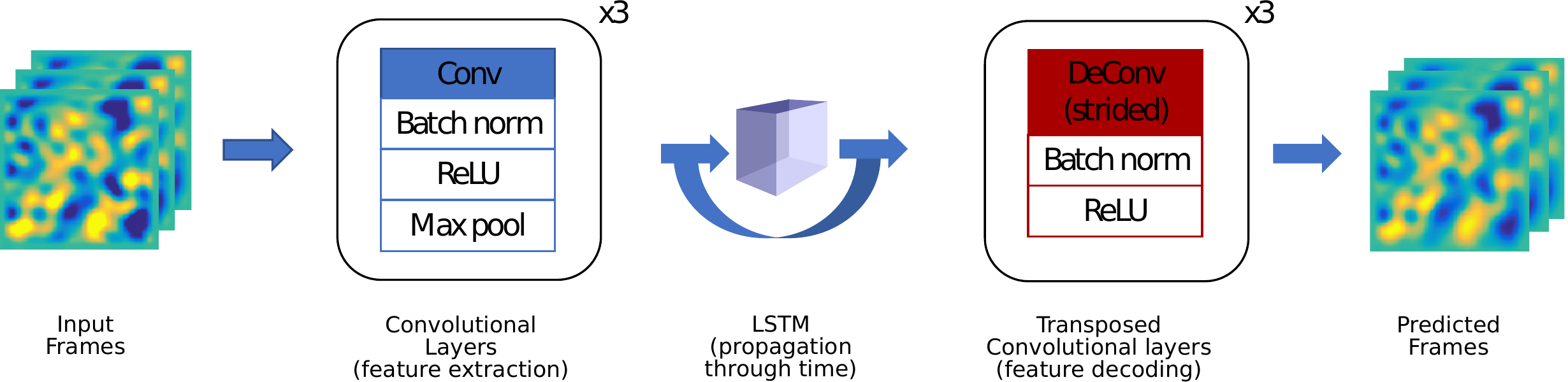}
    \caption{The architecture of the network used to predict the future 
    behaviour of the 2D diffusion system.}
    \label{f:predictionnet}
\end{figure*}

To predict the future behaviour of the system, we built a fully convolutional 
neural network consisting of three main blocks, similar to the architecture 
used by Ehrhardt et al. \cite{ehrhardt_2017}. First, a three-layer encoding 
network extracts relevant features from the input frames, while performing 
dimensionality reduction. By compressing the information that 
passes through the layers, this encoder network thus acts as a bottleneck that 
encourages the network to only extract a useful representation of the system. 
While Ehrhardt et al. \cite{ehrhardt_2017} used a portion of a pre-trained VGG 
network~\cite{simonyan_2014}, here we train the network end-to-end to tailor 
the features extracted to the specific physical system under consideration. 
Next, a convolutional LSTM layer \cite{xingjian_2015} 
progresses the features in time for as many steps as necessary. 
After the recurrent layer, the structure of our predictive network is completed 
by a three-layer decoder network that transforms the output of the LSTM back 
into frames in the spatial domain. Transposed convolutions are used with a 
stride of two \cite{zeiler_2010} to return to the original resolution. Batch 
normalisation \cite{ioffe_2015} and ReLU non-linearities \cite{goodfellow_2016} 
were used after each convolutional layer in both the encoder and the decoder.
A schematic of the neural network can be found in Fig.~\ref{f:predictionnet}, 
while Table~\ref{t:prediction_arch} has a summary of the key network 
parameters. The hyper-parameters of the network were chosen manually, after a 
brief exploration of the hyper-parameter space. Therefore, it is possible that 
a more thorough search would further improve prediction accuracy.

\begin{table}[h]
    \centering
    \begin{tabular}{p{1.5cm}p{1.5cm}p{1.5cm}p{2cm}}
    \toprule
            & Nb. of \newline channels & Filter size & down-/up- \newline 
            sampling \\
    \midrule
    Encoder & (64,32,32)      & (5,5,5)     & Max-pooling (2x2) \\
    LSTM    & 32              & 5           & n.a.              \\
    Decoder & (32,64,K)       & (4,4,4)     & Stride = 2        \\
    \bottomrule
    \end{tabular}
    \caption{Details of the architecture for the prediction network. The number 
    of channels in the last layer of the decoder is variable because it depends 
    on how many frames are requested as output (K).}
    \label{t:prediction_arch}
\end{table}

\subsubsection{Network training}
The network received as input 2 or 3 sequential frames and was trained to 
predict the next 11 frames, while only a smaller number of target frames 
(variable between 1 and 7) was used to back-propagate the prediction error, and 
thereby constituting useful information for the network training process. All 
networks considered in this study were able to extrapolate for more time steps 
than that used during training. We used Mean Squared Error (MSE) -- the average 
of the squared difference between targets and predictions -- as the loss 
function, and trained the network for 1000 epochs using Adam update schemes 
\cite{kinga_2015}, weight decay \cite{ioffe_2015} and a batch size of 64. The 
learning rate started at 0.0005 and was decreased by a factor of ten after 700 
epochs. The simulations were split into training and testing using a 5-fold 
cross-validation scheme. In addition, 20\% of the training dataset was set 
aside for validation, since the final model was chosen as that which exhibited 
best performance across epochs on this validation dataset. The training was 
carried out using the PyTorch framework~\cite{paszke_2017} and required 
between 40 minutes and 60 minutes per network on a GTX 960 GPU (NVIDIA 
Corporation, USA), depending on the number of back-propagated frames.

\subsubsection{Results}

\begin{figure}[tb]
    \centering
    \includegraphics[width=0.9\linewidth]{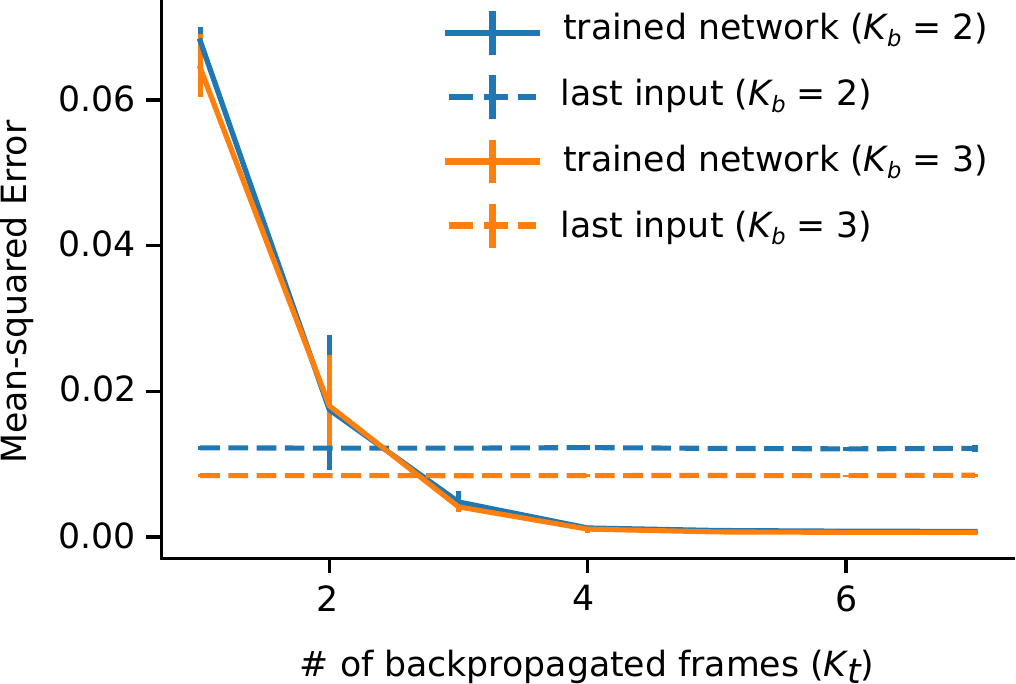}
    \caption{Accuracy of next steps prediction versus number of input ($K_b$) 
        and back-propagated ($K_t$) frames.
        Mean Squared Error (MSE), averaged 
        over the test dataset, first, and the 5 cross-validation folds, then, 
        against $K_t$ and for $K_b=2,3$. Error bars extend to the minimum and 
        maximum MSE among the 5 folds. Dashed lines represent the last input 
        level.}
    \label{f:dnn-predict-frames}
\end{figure}

\begin{figure}[tb]
    \centering
    \includegraphics[width=0.9\linewidth]{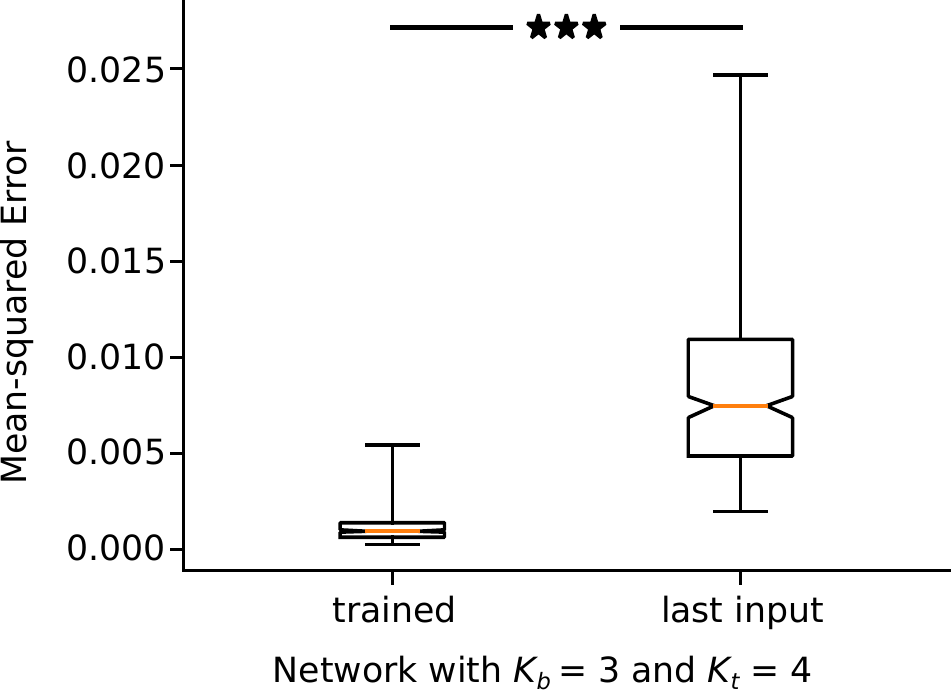}
    \caption{Comparison between the full MSE distribution across the test 
    dataset for the predictions of a single trained network, compared with that 
    given by using the last input frame as the prediction. The asterisks 
    represent statistical 
    significance with p-value $<10^{-4}$ (Wilcoxon signed rank test).}
    \label{f:dnn-predict-comparison}
\end{figure}

\begin{figure}[tb]
    \centering
    \includegraphics[width=0.9\linewidth]{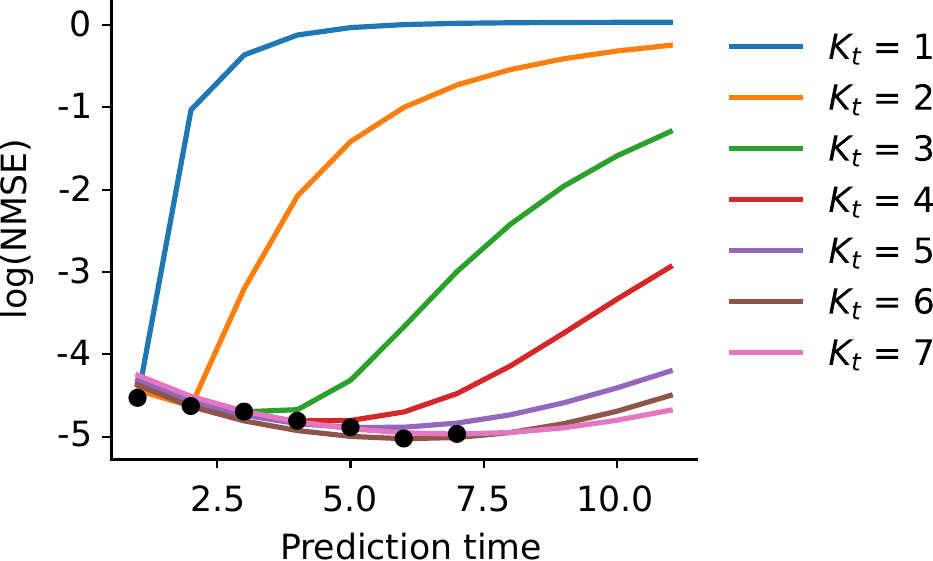}
    \caption{The logarithm of the 
        Normalised Mean Squared Error (NMSE) against the prediction time for 
        networks trained with various $K_t$ (and $K_b=3$). The black dots 
        represent the $K_t$ corresponding to each line.}
    \label{f:dnn-predict-time}
\end{figure}

Our aim was to investigate the changes in prediction accuracy with varying 
amounts of training data, quantified as the number of frames given as the input 
($K_b$) and as the target ($K_t$) to the network. The former determines how 
much information the network can use to make predictions while the latter, 
which corresponds to the number of back-propagated target frames, is used to 
calculate the error which informs the update of the network weights. The more 
frames that are used for back-propagation, the more future time steps can be 
aggregated to build the error signal that guides the network learning process. 
The disadvantage is that longer simulations are required to train the network 
and there is an increased risk of over-fitting.

Prediction errors on the test dataset are shown in 
Fig.~\ref{f:dnn-predict-frames}, for networks trained with different numbers of 
input and target frames. The solid lines correspond to the MSE averaged over 
the test dataset and over the five-fold cross-validation, with the error bars 
extending from the minimum to the maximum accuracy achieved across the five 
folds. The MSE is plotted against the number of back-propagated target frames, 
but is computed as an average across the prediction of 11 future frames, for 
all networks shown. The dashed lines represent the ``last input'' level, 
computed as 
the error that would be achieved if the last input frames were used as the 
predictions. This control test provides a baseline against which to judge the 
generalisation 
capability of the network. Figure~\ref{f:dnn-predict-comparison} shows the full 
``last input'' level distribution against the performance achieved by one of 
the trained 
networks ($K_b=3$, $K_t= 5$). It can be observed that the network is making 
effective use of the information it receives as input, since the accuracy 
achieved is better than that obtained with no physical knowledge for each 
individual simulation in the test dataset.

To assess statistical significance, we compared the two distributions using a 
Wilcoxon signed rank test: we obtained a p-value smaller than $10^{-4}$ against 
the null hypothesis of equality of medians, as shown in 
Figure~\ref{f:dnn-predict-comparison}. Such a null hypothesis can be 
interpreted as the situation in which the neural network makes advantageous use 
of its inputs, but has not learned the physics behind it. Furthermore, the plot 
in 
Fig.~\ref{f:dnn-predict-frames} suggests that the learning capabilities of the 
network saturate after a certain number of back-propagated target frames 
(potentially from $K_t=4$). From that point onwards, the network is able to 
maintain similar error levels when extrapolating the predictions further ahead 
in time. This could be indicative of the network having better assimilated the 
mechanics of diffusion. The benefit of using a smaller number of 
back-propagated frames would be evident when there is a limited time available 
to run the simulations necessary to build the training dataset: in this case, 
being able to train effectively with shorter, rather than longer, simulations 
is advantageous.

It is possible, however, that the gain in prediction accuracy shown in 
Fig~\ref{f:dnn-predict-frames} is influenced by the progressive decrease over 
time of the overall energy of the system, caused by the diffusion dynamics. To 
control for this, we computed the normalised mean squared error (NMSE) -- the 
MSE between predicted and target frames divided by the $L_2$-norm of the target 
frames at each individual time point. Such a measure is similar to the fraction 
of variance explained by a regression algorithm, and allows us to directly 
compare performance at different time steps. Figure~\ref{f:dnn-predict-time} 
shows the logarithm of NMSE as a function of time for networks trained with 
$K_b=3$ and different values of $K_t$, averaged over the test dataset and the 
cross-validation folds. It can be noted that, at each time step, the NMSE is 
inversely proportional to the number of back-propagated frames, suggesting that 
networks with a higher $K_t$ are able to explain a larger proportion of the 
variability of all future frames.

Once trained, each prediction from the neural network took less than one second 
to calculate, compared to 40 seconds required by numerical modelling. Extending 
this to models of cardiac electrophysiology, such a gain in computational speed 
would allow for a quicker exploration of multiple scenarios of interest, such 
as the alteration of functional and structural characteristics of different 
areas of the myocardium when planning intervention.

\subsection{Application to estimating diffusion parameters}
Robustly and accurately estimating the parameters for models is critical for 
them to be useful in prediction. In this section we consider an example of 
how machine learning can meet this need. 
Extending the next-step prediction of Section~\ref{s:pred-dnn}, we explored how 
accurately the parameters of the diffusion model can be estimated from one of 
the networks trained for predictions. We specifically considered the network 
with $K_b=3$ and $K_t=4$, as shown in Fig.~\ref{f:dnn-predict-comparison}.

\begin{table}[tb]
    \centering
    \begin{tabular}{p{2.4cm}p{1.5cm}p{1.5cm}p{1.5cm}}
        \toprule
        & Channels / units & Filter size & down-/up-sampling \\
        \midrule
        Convolutional & (128,64) & (6,6) & Stride = 2 \\
        Fully connected  & (6) & n.a. & n.a. \\
        \bottomrule
    \end{tabular}
    \caption{Details of the architecture for the parameters estimation network.}
    \label{t:dnn-infer-structure}
\end{table}

We first extracted the internal representation of the network for all 11 
predicted future frames. Specifically, this is the activity of the LSTM units. 
This multidimensional vector was used as the input to a second neural network 
with two convolutional, and one fully connected, layers which was trained to 
predict the six parameters $d_0$, $d_1$, $d_{0,scar}$ and $d_{1,scar}$, as well 
as the orientation ($\theta$) and location ($\beta$) of the boundary between 
“healthy” and “scarred” tissue. Details of this network are given in 
Table~\ref{t:dnn-infer-structure}. 

\subsubsection{Results}
\begin{figure}[tb]
    \centering
    \begin{tabular}{p{0.2cm}p{3.5cm}p{0.2cm}p{3.5cm}}
        \vspace{0pt} (a) &
        \vspace{0pt}\includegraphics[width=\linewidth]{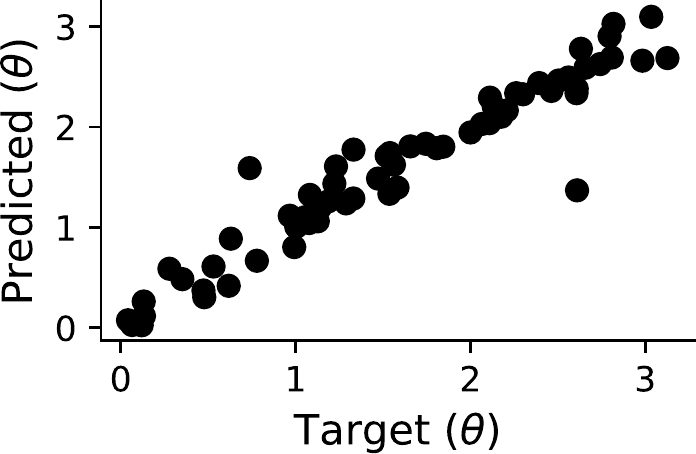}
         &
        \vspace{0pt} (b) &
        \vspace{0pt}\includegraphics[width=\linewidth]{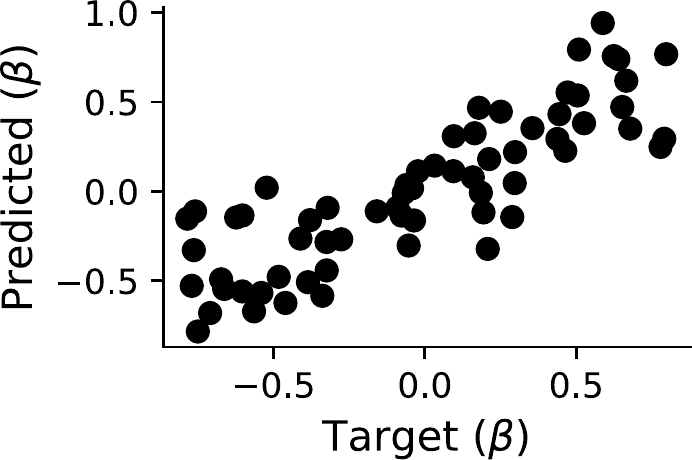}
         \\
        \vspace{0pt} (c) &
        \vspace{0pt}\includegraphics[width=\linewidth]{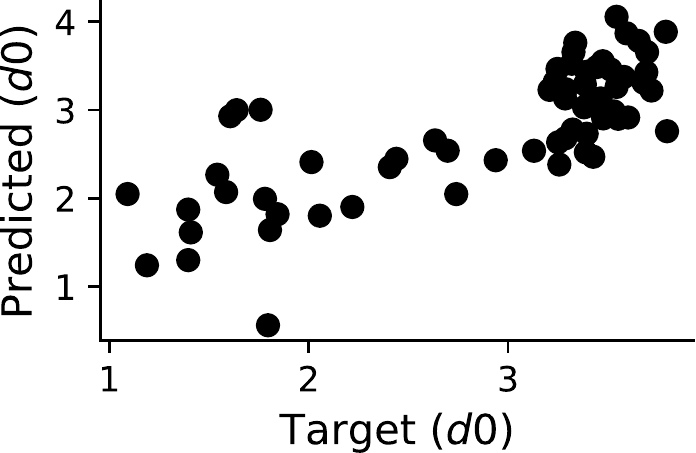}
         &
        \vspace{0pt} (d) &
        \vspace{0pt}\includegraphics[width=\linewidth]{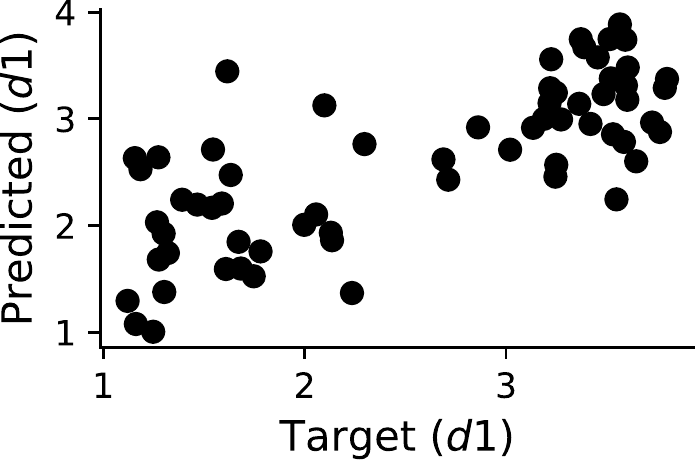}
         \\
        \vspace{0pt} (e) &
        \vspace{0pt}\includegraphics[width=\linewidth]{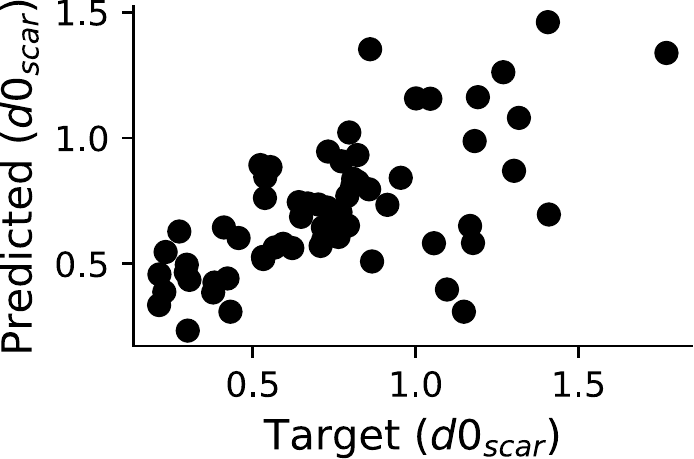}
         &
        \vspace{0pt} (f) &
        \vspace{0pt}\includegraphics[width=\linewidth]{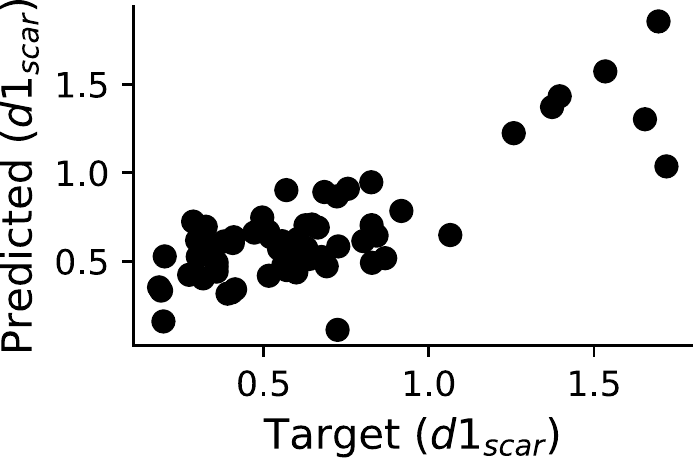}
    \end{tabular}
    
    \caption{Accuracy of parameter inference from the internal representation 
    of the prediction network with $K_t=4$ and $K_b=3$ for (a) boundary angle 
    $\theta$, (b) boundary position $\beta$ and (c-d) scar parameters $d_0$, 
    $d_1$, $d_{0,scar}$ and $d_{1,scar}$. The predicted parameter values are 
    plotted against the target values for a subset of the test dataset.}
    \label{f:dnn-param-inference}
\end{figure}

Results are shown in Fig.~\ref{f:dnn-param-inference}. The correlation 
coefficients between target and predicted parameter values are 0.84, 0.80, 
0.80, 0.70, 0.95, and 0.87 for $d_0$, $d_1$, $d_{0,scar}$, $d_{1,scar}$, 
$\theta$ and $\beta$, respectively. Despite there being potential for 
improvements, the achieved accuracy indicates that the internal representation 
learnt by the network does contain information about physically relevant 
quantities. This suggests that our deep learning model is able to assimilate at 
least some of the mechanisms intrinsic to the physical system under 
consideration.

\section{Approximate Bayesian Computation (ABC)}
\label{s:inference}

Approximate Bayesian Computation (ABC) is a statistical inference technique 
that can be applied as a parameter fitting method to incorporate uncertainty 
estimates into the fitting process \cite{toni_2009,daly_2015,klinger_2017}. The 
algorithm interrogates the range of outputs of a numerical model by drawing 
different parameter choices from a defined \emph{prior} probability space, 
running the simulation, and comparing the output to experimental data. This 
probability space is then sequentially refined based on a distance function 
which measures the closeness of the simulated output to the experimental 
output. After multiple iterations (or reaching a chosen error threshold), the 
algorithm produces a discrete approximation to the true \emph{posterior} 
probability distribution of model parameters given the observed experimental 
data. The distributions give more information to a modeller on the ability of 
the experimental data to constrain the parameter choice and the resulting 
credibility of the full model.

\subsection{Application to inferring cell model parameters}
Tissue-scale cardiac electrophysiology simulations are built on models of the 
action potential of single myocytes. These \emph{cell models} are solved by 
calculating the opening and closing kinetics of transmembrane and internal ion 
currents, and their effect on the membrane potential of the cell. 
Each parameter in ion current sub-models is chosen specific to the particular 
cell type. The parameter values are based on data from patch-clamp experiments 
which interrogate the dynamics of specific ion currents in isolated myocytes at 
a range of prescribed voltages \cite{neher_1992}.

The standard approach is to fit ion channel parameters to these data using a 
traditional method such as least squares regression. These methods produce 
point estimates and thus do not take into account uncertainties introduced 
through the fitting process itself when multiple parameter choices can result 
in similar values of the fitting loss criterion. This has led to discrepancies 
between cell models which purport to represent the same cell type 
\cite{niederer_2009}.

Daly \emph{et al} previously investigated the use of ABC on parameters chosen 
in the original Hodgkin-Huxley action potential model 
\cite{hodgkin_1952,daly_2015}. They found that parameters were generally well 
constrained by the experimental data. These data were average current traces 
which inform both voltaic and temporal behaviour of a channel. Modern 
experimental studies predominantly report steady-state behaviour of channels in 
response to voltage steps.

\subsubsection{Neonatal rat ventricular sodium channel}
We investigate the ability of modern patch-clamp data, which 
may contain less information, to constrain parameters of a physiological model 
for neonatal rat ventricular myocytes (NRVMs) \cite{korhonen_2009}.
We present the result of fitting the fast sodium channel of the NRVM model 
using ABC. The fast sodium channel plays a crucial role in the generation of a 
cellular action potential and the propagation of an electrical signal through 
cardiac tissue; consequently, it is critical to have confidence in any 
\textit{in silico} model of the channel. We use the ABC Sequential Monte Carlo 
(ABCSMC) 
algorithm with a population adaptation strategy from the \emph{pyabc} python 
library (http://pyabc.readthedocs.io/en/latest/) \cite{klinger_2017} and the 
\emph{myokit} python library (https//myokit.org) for running simulations 
\cite{clerx_2016}. The equations for the fast sodium channel 
\cite{korhonen_2009} include three \emph{gates}: activation, fast inactivation, 
and slow inactivation, and are given by,
\begin{align}
I_{Na} &= G_{Na}m^3 h j (V-E_{Na}) \\
m_{\infty} &= \left\{1 + \exp\left[(p_1 + V)/p_2\right]\right\}^{-1}\\
j_{\infty} = h_{\infty} &= \left\{1 + \exp\left[(q_1 + V)/q_2 
\right]\right\}^{-1}\\
\tau_m &= \left\{
\frac{p_3(V+p_4)}{1-\exp\left[p_5(V+p_4\right]}
+ p_6 \exp(-V/p_7)
\right\}^{-1}
\end{align}
where the parameters $p_1$-$p_7$ and $q_1,q_2$ are determined from experimental 
data, the original published values of which are given in the second column of 
Table~\ref{t:korhonen-results}.
The gates govern the proportion of open channels in the cells and thus affect 
the maximum current that is able to flow across the membrane. These equations 
were adapted for NRVMs from earlier cell models of different species by varying 
only the channel conductance \cite{korhonen_2009}. We therefore use only the 
directly applicable data for observations in the ABCSMC algorithm. These data 
are from patch clamp experiments on adult rat ventricular cells 
\cite{pandit_2001,lee_1999}. They include five patch clamp protocols testing 
activation, inactivation and recovery characteristics of the channel. The 
protocols do not explicitly test temporal characteristics of the current, and 
thus we retain the temporal parameters of the fast and slow inactivation 
processes to reduce the dimensionality of the problem. For each of the nine 
parameters that we constrain using ABCSMC, initial priors were set to uniform 
distributions roughly an order of magnitude larger than the parameter setting 
in the original model.

\begin{table}[tb]
    \centering
    \begin{tabular}{llllll}
    \toprule
    & orig. & prior     & posterior & & \\
    \cmidrule{4-6}
    &&& mean    & min    & max     \\
    \midrule
p1        & 45       & (0, 100)  & 43.4    & 43.2    & 43.6    \\
p2        & -6.5     & (-50, 0)  & -11.6   & -11.6   & -11.5   \\
p3        & 0.235    & (0, 1)    & 0.0717  & 0.0707  & 0.0726  \\
p4        & 47.1     & (0, 100)  & 79.7    & 79.3    & 79.9    \\
p5        & -0.1     & (-50, 0)  & -36.2   & -50.0   & -12.2   \\
p6        & 0.0588   & (0, 1)    & 0.00345 & 0.00302 & 0.00373 \\
p7        & 11.0     & (0, 1000) & 682     & 325     & 998     \\
q1        & 76.1     & (0, 100)  & 72.5    & 72.9    & 73.1    \\
q2        & 6.07     & (0, 50)   & 9.54    & 8.82    & 10.0    \\
    \bottomrule
    \end{tabular}
    \caption{Results of ABCSMC inference for the parameters of the fast sodium 
    channel.}
    \label{t:korhonen-results}
\end{table}

\subsubsection{Results}

Table~\ref{t:korhonen-results} also shows the prior distribution ranges and 
statistics of the posterior distributions. Seven of the nine parameters appear 
well constrained by the data, of which four ($p_1$, $p_2$, $q_1$, $q_2$) are 
relatively close to the original values. These parameters govern the 
steady-state activation and inactivation (both fast and slow) of the current, 
confirming that the patch clamp protocols predominantly test the steady-state 
characteristics of this current.

\begin{figure*}[tb]
    \centering
    \begin{tabular}{p{0.3cm}p{17.5cm}}
    \vspace{0pt}
    (a) &
    \vspace{0pt}
    \includegraphics[width=\linewidth]{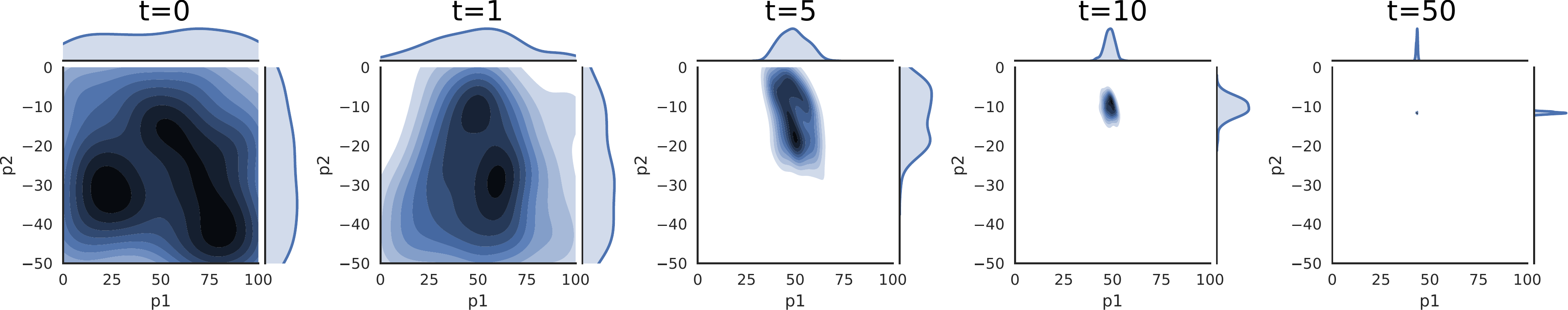}\\
    \vspace{0pt}
    (b) &
    \vspace{0pt}
    \includegraphics[width=\linewidth]{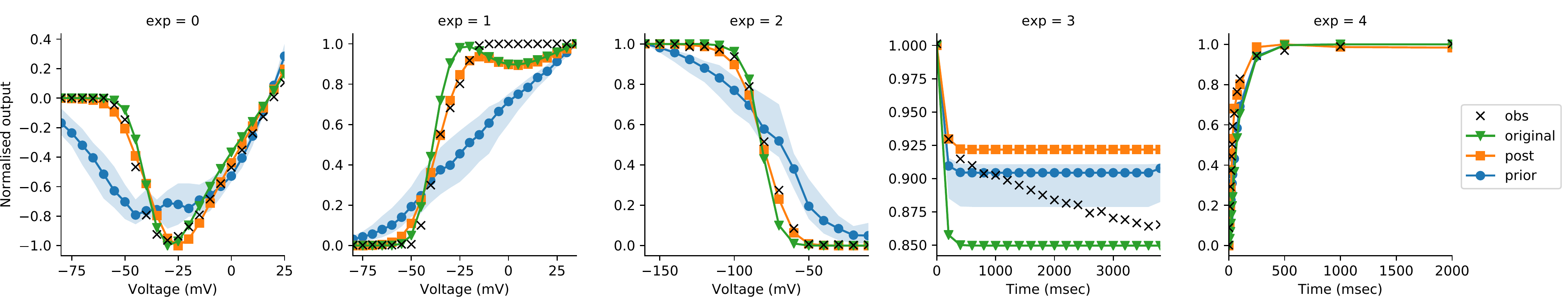}
    \end{tabular}
    \caption{(a) Kernel density estimates for steady-state activation 
    parameters showing sequential constraining of distributions. (b) Data used 
    to fit channel (black crosses) are plotted for each patch clamp experiment. 
    Simulation results are also plotted for original parameter settings (green 
    triangles), 100 samples from prior distribution (blue circles), and 100 
    samples from posterior distribution (orange squares). Shaded area indicates 
    95\% confidence intervals around the median line.}
    \label{f:korhonen}
\end{figure*}

Figure~\ref{f:korhonen}(a) shows how the distributions of the two steady-state 
parameters of activation are sequentially constrained through the iterations of 
ABCSMC. Figure~\ref{f:korhonen}(b) shows the data used to fit the current, 
along with simulation results using original parameters, 100 prior distribution 
samples and 100 posterior distribution samples. The output for the posterior 
distribution is close to both the original settings and the experimental data. 
In some aspects, particularly the upper half of inactivation behaviour in 
exp=2, the ABCSMC fit is a noticeable improvement over the original parameter 
choices. However, in other areas such as the start of activation (seen in exp=0 
and exp=1), the ABCSMC fit is further from the observed data than the original 
settings. For exp=3 which tests the normalised peak current of a regular train 
of voltage pulses, the equations of the model appear unable to capture both the 
initial exponential and then linear decay of the observed data, shown by the 
fact that both the ABC posterior and original parameter choices end in a 
constant relationship after the initial decay portion.

Despite the large variation present in two of the parameters, the posterior 
results in Figure~\ref{f:korhonen}(b) show little variation. This indicates 
that the current patch-clamp protocols may not sufficiently interrogate 
temporal aspects of the channel, as both unconstrained parameters (p5 and p7) 
govern this aspect of the model equations. This highlights the value of the ABC 
approach; using traditional fitting methods, we would not be aware of the 
unidentifiability in these parameters. More complex patch-clamp protocols could 
be investigated in an attempt to improve the ability of the data to constrain 
the model.

\section{Discussion}
\label{s:discussion}

In this review we have shown the potential benefits of a number of 
machine learning approaches and how they can 
enable us to extract more information from the data we collect. The electrogram 
is the ubiquitous data modality of the cardiac electrophysiology catheter 
laboratory and yet the relative information content extracted from these 
complex signals is currently poor. We have demonstrated that by quantifying and 
combining a range of features in the signal, some of which are already used in 
isolation, and applying machine learning algorithms to them we can learn more 
about the properties of the underlying myocardium and the substrate that 
sustains arrhythmias. Deep learning allows us to further automate this process, 
removing the inherent bias of manually choosing potentially sub-optimal 
features, and allowing the neural network to extract latent representations 
which best discriminate between classes directly from the signal.

Numerical modelling is becoming increasingly established within the cardiac electrophysiology field, due to the increased availability of computational power, and improved resolution of clinical imaging technologies. However, the numerical resolution requirements for action potential propagation and complexity of ion channel kinetics still necessitates high computational cost. Furthermore, the number of parameters and difficulties associated with deriving appropriate values experimentally or clinically means that great care is needed when incorporating these into predictive modelling. We have shown how machine learning can help in both inferring appropriate parameter values from data as well as quantifying how certain we can be in those parameter values and therefore how confident we can be in the model output.

\subsection{Limitations}
In analysing the micro-electrode array electrogram data in Section~\ref{s:egm} 
the large-amplitude stimulus artefact, created by pacing of the culture, was 
first 
removed. This was required for the feature-detection algorithms to reliably 
measure the electrogram characteristics used to form the 
feature vector. Since the stimulus artefact dominates the signal, it was also a 
necessary pre-processing step for the convolutional neural network approach. 
Without its removal the network was unable to distinguish specific 
morphological features of the response signal.

For the robust training of deep neural networks, used in both 
Section~\ref{s:cnn} and Section~\ref{s:rnn}, the volume of data and 
computational cost of training is high. While data augmentation techniques are 
used to improve the generalisation of models, additional recorded data would 
further improve the quality of the predictions made by these methods.

Graphics Processing Units (GPUs) are particularly effective at undertaking the 
learning process for deep neural networks and their use is essential to produce 
trained models in tractable timescales. In this context, feature-based 
classifiers provide a performance advantage in situations where appropriate 
features are known and can be defined \emph{a priori} to distinguish the 
classes. However, the computational time for prediction using the trained 
models is negligible for both feature-based and deep-learning based methods.

\section{Conclusions}
\label{s:conclusions}
Cardiac arrhythmias are a major global healthcare problem and there is 
significant scope for improving their diagnosis and treatment. Improvements 
will be achieved from better understanding of the mechanisms sustaining 
fibrillation, as well as increasingly personalised treatment. Modern machine 
learning techniques and numerical modelling, when applied appropriately, both 
have great potential to help fulfil this role and their combination, in 
particular, offers a powerful approach to achieving personalisation of care.




\section*{Acknowledgements}
This work was generously supported by the Rosetrees Trust (grant M577), British Heart Foundation research grants (PG/15/59/31621, PG/16/17/32069 and RG/16/3/32175) and the British Heart Foundation Centre for Research Excellence (grant RE/13/4/30184).

\bibliographystyle{elsarticle-num-names}
\bibliography{cep-rethinking-ml.bib}







\end{document}